\documentclass[10pt,twocolumn,letterpaper]{article}

\usepackage{cvpr} 
\usepackage{times}
\usepackage{epsfig}
\usepackage{graphicx}
\usepackage{amsmath}
\usepackage{amssymb}

\usepackage{multirow}
\usepackage{booktabs}
\usepackage[pagebackref=true,breaklinks=true,letterpaper=true,colorlinks,bookmarks=false]{hyperref}

\begin{document}

\title{MAGE-Vein: Multi-Instance Age and Gender Estimation from Finger Vein Images}

\author{Katsuki Tanaka$^1$, Koichi Ito$^1$, Takafumi Aoki$^1$, Masakazu Fujio$^2$,Yosuke Kaga$^2$, \\Kanade Oshima$^2$, and Kenta Takahashi$^2$,\\
1 Graduate School of Information Sciences, Tohoku University, Japan.\\
2 Hitachi, Ltd., Japan.\\
{\tt\small tanaka@aoki.ecei.tohoku.ac.jp}
}

\maketitle
\thispagestyle{empty}

\begin{abstract}
  Age estimation from finger vein images has been widely considered impractical due to severe demographic biases in public datasets and physiological confounding factors like gender.
  To overcome these limitations, we propose MAGE-Vein, a novel multi-instance, multi-task learning framework.
  Our approach extracts robust structural aging signs by employing a hybrid feature-level fusion of three fingers, effectively suppressing local imaging noise.
  Furthermore, simultaneous optimization of gender classification conditions the network to effectively eliminate gender-specific vascular variations.
  Evaluated on a demographically balanced dataset of 402 subjects, MAGE-Vein achieves a mean absolute error of 6.12 years and a correlation of 0.880.
  Our results not only overturn the conventional consensus regarding the limitations of the finger vein modality but also demonstrate that previous estimation failures were primarily artifacts of biased public datasets.
  Our code is available at \url{https://github.com/gsisaoki/MAGE-Vein}.
\end{abstract}

\section{Introduction}

Finger vein recognition has been widely deployed in financial transactions and access control as a biometric technology offering high spoofing resistance and user convenience \cite{Handbook-Biometrics}. 
While finger vein patterns are assumed to remain invariant throughout a person's lifetime, vein characteristics actually change with age \cite{Huang-ClinImaging-2021}.
Consequently, template aging has emerged as a critical issue \cite{Hall-ICISSP-2022}.
This aging phenomenon is driven by physiological changes such as reduced peripheral blood flow, increased vascular tortuosity, and skin thinning.
Indeed, our analysis of the collected dataset (detailed later) demonstrated these changes; the average vessel thickness slightly increases from 2.40 pixels for subjects in their 10s to 2.46 pixels for those in their 70s, while the average image brightness rises from 85.81 to 86.49.
To mitigate the degradation of matching accuracy caused by such aging effects and to enhance system robustness, it is necessary to estimate soft biometric attributes, particularly age, directly from finger vein images.

To date, attribute estimation has been explored across various biometric modalities, including faces \cite{Shou-Pattern-2025,Kumar-Sensors-2022}, fingerprints \cite{Falohun-Appl-2016}, irises \cite{Gowroju-Multimed-2022,Costa-Abreu-Biosig-2015}, and palm veins \cite{Hernandez-ICPRS-2022}.
In the context of finger veins, Convolutional Neural Networks (CNNs) have been employed for both gender estimation \cite{Kuzu-Access-2023} and age estimation \cite{Wimmer-ICPRW-2023}.
However, Wimmer et al. \cite{Wimmer-ICPRW-2023} concluded that age estimation from finger veins is highly challenging and impractical.
Based on our analysis, this negative conclusion stems from the significant demographic bias in existing public datasets, such as MMCBNU \cite{Lu-CISP-2013} and UTFVP \cite{Ton-ICB-2013}.
Since the samples in these datasets are concentrated in young and middle-aged demographic groups, age estimation models tend to overfit to the majority classes.
Furthermore, anatomical differences between genders, such as variations in vessel diameter and hemoglobin concentration \cite{Kuzu-Access-2023}, serve as a confounding factor in age estimation.
Since conventional methods treat age estimation as a single task, they cannot separate these variations and extract age-related structural changes.

To address these issues, this paper proposes MAGE-Vein (Multi-instance Age and Gender Estimation from finger Vein images), a multi-task learning framework designed to achieve accurate age estimation from finger vein images.
Specifically, our method employs a multi-instance fusion strategy that integrates feature representations from three fingers (index, middle, and ring fingers) of the same subject at the feature level.
This approach effectively smooths out local imaging noise caused by variations in finger pressure or blood flow, enabling the stable extraction of global age-related vascular changes common to the entire vascular system.
Furthermore, the framework simultaneously optimizes age estimation and gender classification using a shared backbone network.
This multi-task strategy conditions the network on gender-specific vascular traits, such as thickness and contrast, making it possible to separate and extract structural changes associated with aging.
Through comprehensive experiments utilizing a demographically balanced dataset of 402 subjects, the proposed method achieves a Mean Absolute Error (MAE) of 6.12 years and a correlation of 0.880.
These results demonstrate the effectiveness of MAGE-Vein in age estimation from finger vein images.

\section{Related Work}
\label{sec:related_work}

In the context of attribute estimation from finger vein images, particularly gender classification, several studies have proposed highly accurate methods by introducing deep learning techniques.
Kuzu et al. \cite{Kuzu-Access-2023} proposed a gender classification method by resizing images to 224$\times$224 pixels and extracting features using DenseNet-161 \cite{Hung-CVPR-2017}.
Furthermore, Wimmer et al. \cite{Wimmer-ICPRW-2023} proposed methods utilizing DenseNet-161 and SqueezeNet \cite{Iandola-ICLR-2013} trained with a triplet loss, demonstrating that finger vein images contain sufficient discriminative features for gender classification.

\begin{figure}[t]
  \centering
  \includegraphics[width=.84\linewidth]{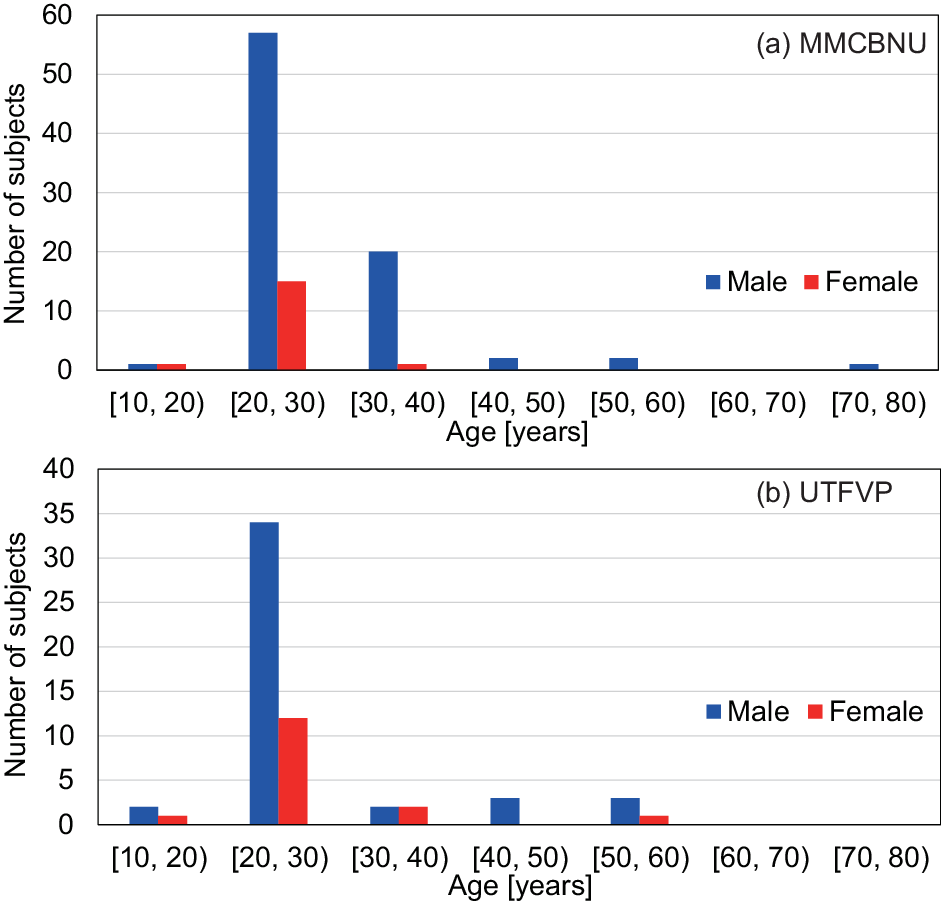}
  \caption{Age and gender distributions of public finger vein datasets: (a) MMCBNU \cite{Lu-CISP-2013} and (b) UTFVP \cite{Ton-ICB-2013}.
  The concentration of subjects in 20s and 30s in both datasets illustrates a significant demographic bias.
  This fact limits the effective learning of age-related features, causing models to overfit to the mean age.}
  \label{fig:public_db}
\end{figure}

However, age estimation from finger vein images remains an unresolved challenge.
Although Wimmer et al. \cite{Wimmer-ICPRW-2023} attempted age regression using ResNeXt-101 \cite{Xie-cvpr-2017}, they concluded that age estimation from this modality is highly challenging and impractical.
Based on our analysis, this negative outcome primarily stems from the significant demographic imbalances in the public datasets rather than algorithmic limitations.
Fig. \ref{fig:public_db} illustrates the age and gender distributions of MMCBNU \cite{Lu-CISP-2013} and UTFVP \cite{Ton-ICB-2013}, the public datasets evaluated by Wimmer et al.
These datasets exhibit demographic skewness, with over 80\% of the subjects concentrated in the 20- and 30-year-old age groups.
When trained on such imbalanced data, age estimation models fail to capture diverse age-related structural changes and overfit to predict values near the mean age of the dataset, resulting in a loss of linearity in age regression.

Furthermore, the precision of age labels in existing datasets remains an additional challenge.
Datasets such as MMCBNU provide age labels only as integer values.
Because these integer labels do not account for exact dates of birth, they introduce fractional uncertainties.
This imprecision reduces their reliability as ground truth values for regression tasks, thereby contributing to increased estimation errors.
To address these limitations, this paper utilizes a demographically balanced dataset of 402 subjects featuring highly precise chronological age labels calculated in fractional years.
By employing this dataset, we aim to clarify the age-related structural changes inherently present in finger vein images.

\begin{figure*}[t]
  \centering
  \includegraphics[width=.83\linewidth]{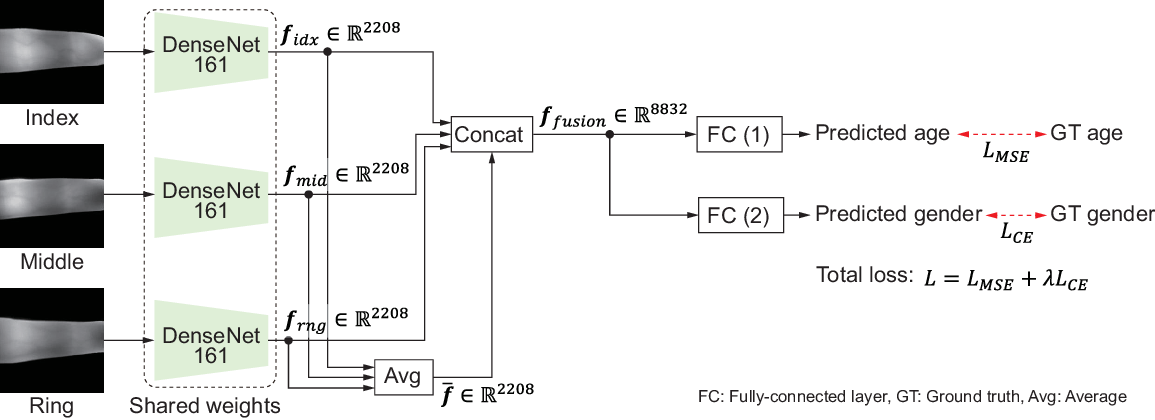}
  \caption{Overview of the proposed MAGE-Vein framework.
  Our multi-instance approach constructs a robust feature representation $\mathbf{f}_{fusion}$ by fusing concatenated and averaged features from three fingers (index, middle, and ring), effectively suppressing local imaging noise.
  Furthermore, simultaneous optimization of age regression ($L_{MSE}$) and gender classification ($L_{CE}$) isolates structural aging signs by conditioning the network on gender-specific vascular variations.}
  \label{fig:proposed}
\end{figure*}

\section{Multi-Instance Age and Gender Estimation}

This section describes MAGE-Vein, a multi-instance, multi-task learning framework designed to jointly estimate age and gender from finger vein images.
The overview of MAGE-Vein is illustrated in Fig. \ref{fig:proposed}.
The proposed method achieves accurate age estimation by simultaneously considering the anatomical correlations among multiple fingers and the gender-specific variations in vascular traits, rather than relying on the limited information from a single finger.

\paragraph{Multi-Instance Feature Extraction}

During the acquisition of finger vein images, variations in finger pressure, transient fluctuations in blood flow, and unstable contact angles with the device introduce local noise to the contrast and thickness of the blood vessels.
Since relying on a single finger input carries the risk of misinterpreting such noise as age-related changes, the proposed method takes three instances from the same subject, i.e., the index, middle, and ring fingers, as input.
For the backbone, we adopt DenseNet-161 \cite{Hung-CVPR-2017}, which efficiently extracts high-dimensional residual features and achieved the highest estimation accuracy in our preliminary experiments.
The three fingers are fed into the shared backbone network.
This allows the network to acquire global feature representations common to the entire vascular system of the subject while preserving the specific texture of each finger.

\paragraph{Feature Fusion Strategy}

When integrating the extracted features $f_{idx}, f_{mid}, f_{rng} \in \mathbb{R}^{2208}$ from each finger, the proposed method employs a hybrid fusion strategy that combines concatenation with statistical averaging.
Specifically, we compute the concatenated vector of the individual features from the three fingers, $\mathbf{f}_{concat} = [f_{idx}, f_{mid}, f_{rng}]$, and their average feature, $\bar{f} = \frac{1}{3} \sum f_i$, to construct the final fused feature vector $\mathbf{f}_{fusion} = [\mathbf{f}_{concat}, \bar{f}] \in \mathbb{R}^{8832}$.
This design enhances robustness against fluctuations in the imaging environment by prompting the network to simultaneously reference the unique structural changes of each finger and the individual-level anatomical biases.

\paragraph{Multiple Attribute Estimation}

The constructed fused feature $\mathbf{f}_{fusion}$ is fed into two linear layers: an age estimation head (regression) and a gender classification head (classification).
Significant gender differences exist in vessel diameter and hemoglobin concentration in finger veins, which serve as a confounding factor in age estimation.
By simultaneously optimizing the gender classification task, the proposed method conditions the network on gender-specific vascular traits, enabling it to separate and utilize morphological and topological changes associated with aging for age estimation.

\paragraph{Loss Function}

To train the model, we employ a combined loss function $L$ that integrates the Mean Squared Error (MSE) loss to capture the continuous nature of aging and the Cross-Entropy (CE) loss to capture the discrete nature of gender, defined by
\begin{equation}
  L = L_{MSE} + \lambda L_{CE},
\end{equation}
where $\lambda$ is a hyperparameter that controls the contribution of the gender information.

\section{Experiments and Discussion}

This section evaluates the effectiveness of the proposed multi-instance, multi-task learning framework, MAGE-Vein.
The experiments begin with the selection of the optimal backbone network and preprocessing method.
Subsequently, we conduct ablation studies to systematically investigate the impact of multi-instance fusion and multi-task learning on the estimation accuracy.
Finally, we demonstrate the advantages of the proposed method in age and gender estimation from finger vein images through quantitative comparisons with conventional methods and qualitative analysis using Grad-CAM.

\begin{table*}[t]
  \centering
  \caption{Comparison of backbone architectures for the proposed method. Best results are in bold, and second-best are underlined.}
  \label{tab:backbone}
  \begin{tabular}{lcccc}
    \toprule
    Backbone & MAE [y/o] $\downarrow$ & Corr. $\uparrow$ & CS@5 $\uparrow$ & Std. $\downarrow$ \\
    \midrule
    ResNet-50 \cite{He-cvpr-2016} & 8.81 & 0.735 & 0.330 & 10.90 \\
    ResNet-101 \cite{He-cvpr-2016} & 9.42 & 0.697 & 0.319 & 11.57 \\
    ResNet-152 \cite{He-cvpr-2016} & 9.40 & 0.704 & 0.319 & 11.42 \\
    DenseNet-121 \cite{Hung-CVPR-2017} & \underline{8.79} & 0.738 & 0.360 & 10.85 \\
    DenseNet-161 \cite{Hung-CVPR-2017} & 8.81 & \underline{0.741} & \underline{0.362} & \underline{10.82} \\
    DenseNet-169 \cite{Hung-CVPR-2017} & 9.59 & 0.682 & 0.316 & 11.89 \\
    DenseNet-201 \cite{Hung-CVPR-2017} & 9.89 & 0.737 & 0.316 & 10.91 \\
    ResNeXt-50-32x4d \cite{Xie-cvpr-2017} & \textbf{8.59} & \textbf{0.745} & \textbf{0.368} & \textbf{10.76} \\
    ResNeXt-101-32x8d \cite{Xie-cvpr-2017} & 9.38 & 0.697 & 0.319 & 11.57 \\
    \midrule
    ViT-B \cite{Dosovitskiy-ICLR-2021} & 13.56 & -0.194 & 0.184 & 16.08 \\
    ConvNeXt-B \cite{Liu-cvpr-2022}  & 10.58 & 0.585 & 0.296 & 13.13 \\
    DINOv2 ViT-B/14 \cite{Oquab-TMLR-2024} & 12.12 & 0.422 & 0.224 & 14.69 \\
    DINOv2 ViT-B/14 with Reg. \cite{Oquab-TMLR-2024} & 12.04 & 0.450 & 0.222 & 14.51 \\
    \bottomrule
  \end{tabular}
\end{table*}

\subsection{Datasets and Experimental Conditions}

This subsection describes the details of the datasets used for evaluation, the implementation conditions for training the proposed method, and the evaluation metrics for estimation accuracy.

\paragraph{Datasets and Preprocessing}

\begin{figure}[t]
  \centering
  \includegraphics[width=.84\linewidth]{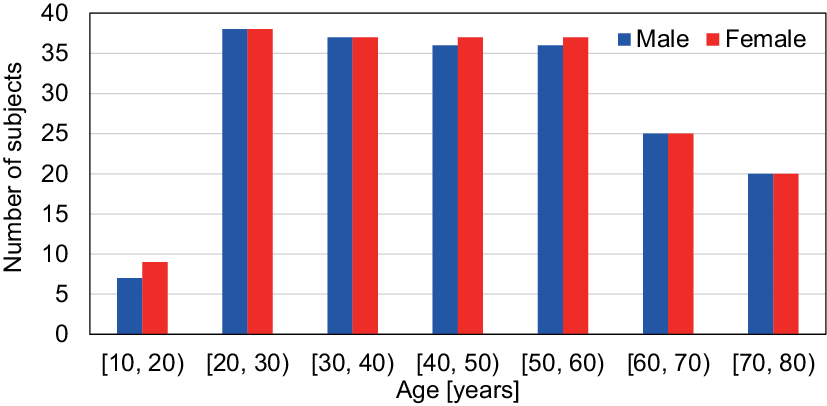}
  \caption{Age and gender distributions of subjects in our dataset.}
  \label{fig:dataset}
\end{figure}

To address the issue of demographic bias pointed out in existing studies \cite{Hall-ICISSP-2022}, we collected and utilized a large-scale finger vein dataset comprising 402 subjects with balanced age and gender distributions.
The subjects consist of 199 males and 203 females ranging in age from their 10s to 70s.
Images of six fingers in total, i.e., the index, middle, and ring fingers of both hands, were captured from each subject using a reflective near-infrared device.
The detailed demographic distribution of the subjects is illustrated in Fig. \ref{fig:dataset}.
To eliminate label uncertainties in regression learning, highly precise chronological ages calculated in fractional years based on birth dates were adopted as age labels.
In the preprocessing stage, to avoid the effects of structural distortion caused by resizing on feature extraction \cite{Lu-Sensors-2013}, all images were normlized to a 244$\times$244-pixel square format using padding.
We utilize a total of 31,575 images: 279 subjects (21,966 images) for training, 79 subjects (6,192 images) for validation, and 44 subjects (3,417 images) for testing.
The collection of this dataset was ethically and legally conducted with the approval of our Institutional Review Board (IRB), and prior informed consent was obtained from all subjects.
This dataset remains private due to legal and privacy protection considerations concerning sensitive biometric data.

\paragraph{Implementation Details}

For model training, AdamW \cite{Loshchilov-ICLR-2019} was used as the optimizer, with the batch size set to 16, the initial learning rate to 0.0001, and the weight decay rate to 0.01.
To suppress overfitting, early stopping was introduced, which terminates the training if the validation loss does not improve for 10 consecutive epochs.
The weights at the minimum validation loss were used for the final evaluation.
For data augmentation, considering the fluctuations during image acquisition, horizontal flipping was applied simultaneously to the three fingers with a probability of 40\%.

\paragraph{Evaluation Metrics}

As performance metrics for age estimation, we employ Mean Absolute Error (MAE), the correlation coefficient (Corr.), and the Cumulative Score at 5 years (CS@5), which represents the accuracy rate within an error margin of $\pm$5 years.
Furthermore, to evaluate the stability of the model, the standard deviation (Std.) of the individual estimation errors is calculated.
For gender classification, based on the importance of considering gender-specific variations in vascular traits \cite{Kuzu-Access-2023}, the F1-score is adopted as the comprehensive evaluation metric.

\begin{table*}[t]
  \centering
  \caption{Impact of input image preprocessing (padding vs. resize) on age estimation accuracy. Best results for each model are in bold.}
  \label{tab:preprocess}
  \begin{tabular}{llcccc}
    \toprule
    Backbone & Preprocess & MAE [y/o] $\downarrow$ & Corr. $\uparrow$ & CS@5 $\uparrow$ & Std. $\downarrow$ \\
    \midrule
    \multirow{2}{*}{DenseNet-121 \cite{Hung-CVPR-2017}} 
     & Padding & \textbf{8.79} & \textbf{0.738} & \textbf{0.360} & \textbf{10.85} \\
     & Resize  & 8.86 & 0.735 & 0.341 & 10.92 \\
    \midrule
    \multirow{2}{*}{DenseNet-161 \cite{Hung-CVPR-2017}} 
     & Padding & 8.81 & 0.741 & \textbf{0.362} & 10.82 \\
     & Resize  & \textbf{8.72} & \textbf{0.744} & 0.344 & \textbf{10.79} \\
    \midrule
    \multirow{2}{*}{ResNeXt-50-32x4d \cite{Xie-cvpr-2017}} 
     & Padding & \textbf{8.59} & \textbf{0.745} & \textbf{0.368} & \textbf{10.76} \\
     & Resize  & 8.87 & 0.731 & 0.340 & 11.09 \\
    \midrule
    \multirow{2}{*}{ResNeXt-101-32x8d \cite{Xie-cvpr-2017}} 
     & Padding & 9.38 & 0.697 & \textbf{0.319} & 11.57 \\
     & Resize  & \textbf{9.31} & \textbf{0.728} & 0.302 & \textbf{11.07} \\
    \bottomrule
  \end{tabular}
\end{table*}

\subsection{Ablation Study}

This section conducts the ablation study to validate the components of MAGE-Vein.

\paragraph{Ex. 1: Selection of Backbone Architecture and Preprocessing}

First, we evaluated various backbones for age estimation using single-finger inputs, as shown in Table \ref{tab:backbone}.
Vision Transformers \cite{Dosovitskiy-ICLR-2021} and DINOv2 \cite{Oquab-TMLR-2024} performed poorly, confirming the superiority of CNNs.
This result indicates that the inductive bias of CNNs is better suited for extracting local textures, such as minute vessel tortuosity and thickness, from finger vein images lacking global semantic context.
We selected the top four models, i.e., DenseNet-121 \cite{Hung-CVPR-2017}, DenseNet-161 \cite{Hung-CVPR-2017}, ResNeXt-50-32x4d \cite{Xie-cvpr-2017}, and ResNeXt-101-32x8d \cite{Xie-cvpr-2017}, as baselines.
Next, we investigated the impact of preprocessing using these baselines.
Conventional methods \cite{Kuzu-Access-2023,Wimmer-ICPRW-2023} employed resizing to normalize the image size, resulting in the distortion of vascular topology and finger contour aspect ratios, and destroying age-related anatomical features.
Table \ref{tab:preprocess} compares padding that preserves the aspect ratio (244$\times$244 pixels) with conventional resizing (224$\times$224 pixels).
Although resizing slightly improved the MAE for certain models like DenseNet-161, padding generally achieved comparable or higher accuracy (e.g., an MAE of 8.59 for ResNeXt-50) while reducing the standard deviation (Std.), leading to more stable estimations.
Consequently, to prevent anatomical distortion, we adopt padding as a preprocessing method.

\paragraph{Ex. 2: Effectiveness of Multi-Instance Input}

Table \ref{tab:multi_instance} compares the age estimation accuracy among three approaches: single-finger input, score-level fusion (averaging independent age predictions from three fingers), and the proposed feature-level fusion (integrating features within the network to output a single estimation).
Feature-level fusion consistently achieved the highest accuracy across all backbones.
Single-finger inputs tend to yield unstable estimations since they are directly susceptible to local noise caused by minute variations in finger pressure and contact angles.
Furthermore, score-level fusion, which merely averages the final prediction scores, yielded limited accuracy improvements.
In models such as ResNeXt-101, performance degradation indicating training collapse was even observed.
In contrast, the proposed feature-level fusion contributes to substantial and stable accuracy improvements.
This result suggests that integrating features internally enables the network to complementarily extract universal age-related features common across the three fingers.

\begin{table*}[t]
  \centering
  \caption{Comparison of multi-instance input and fusion methods. Best results for each model are in bold, and second-best are underlined.}
  \label{tab:multi_instance}
  \begin{tabular}{lllcccc}
    \toprule
    Backbone & Modality & Fusion Strategy & MAE [y/o] $\downarrow$ & Corr. $\uparrow$ & CS@5 $\uparrow$ & Std. $\downarrow$ \\
    \midrule
    \multirow{5}{*}{DenseNet-121 \cite{Hung-CVPR-2017}}
     & \multirow{3}{*}{Single-instance} 
      & Index only  & \underline{7.83} & 0.791 & \underline{0.393} & 9.85 \\
     && Middle only & 9.82 & 0.665 & 0.288 & 12.01 \\
     && Ring only   & 8.12 & 0.783 & 0.378 & 10.01 \\
    \cmidrule{2-7}
     & \multirow{2}{*}{Multi-instance}  
      & Score-level fusion & 7.45 & \underline{0.851} & 0.349 & \underline{8.53} \\
     && Feature-level fusion (Ours) & \textbf{6.76} & \textbf{0.864} & \textbf{0.483} & \textbf{8.25} \\
    \midrule
    \multirow{5}{*}{DenseNet-161 \cite{Hung-CVPR-2017}}
     & \multirow{3}{*}{Single-instance} 
      & Index only  & 8.89 & 0.744 & 0.306 & 10.77 \\
     && Middle only & 8.86 & 0.738 & 0.347 & 10.89 \\
     && Ring only   & 8.43 & 0.772 & 0.329 & 10.23 \\
    \cmidrule{2-7}
     & \multirow{2}{*}{Multi-instance}  
      & Score-level fusion & \underline{8.06} & \underline{0.837} & \textbf{0.431} & \underline{8.95} \\
     && Feature-level fusion (Ours) & \textbf{7.25} & \textbf{0.840} & \underline{0.415} & \textbf{8.83} \\
    \midrule
    \multirow{5}{*}{ResNeXt-50-32x4d \cite{Xie-cvpr-2017}}
     & \multirow{3}{*}{Single-instance} 
      & Index only  & 9.19 & 0.727 & 0.291 & 11.05 \\
     && Middle only & 9.59 & 0.675 & 0.317 & 11.88 \\
     && Ring only   & 8.94 & 0.721 & 0.344 & 11.16 \\
    \cmidrule{2-7}
     & \multirow{2}{*}{Multi-instance}  
      & Score-level fusion & \underline{8.12} & \underline{0.808} & \textbf{0.436} & \underline{9.65} \\
     && Feature-level fusion (Ours) & \textbf{7.40} & \textbf{0.823} & \underline{0.402} & \textbf{9.14} \\
    \midrule
    \multirow{5}{*}{ResNeXt-101-32x8d \cite{Xie-cvpr-2017}}
     & \multirow{3}{*}{Single-instance} 
      & Index only  & 9.35 & 0.755 & 0.357 & 10.56 \\
     && Middle only & 9.46 & 0.697 & 0.301 & 11.55 \\
     && Ring only   & \underline{8.39} & 0.750 & \textbf{0.390} & 10.67 \\
    \cmidrule{2-7}
     & \multirow{2}{*}{Multi-instance}  
      & Score-level fusion & 14.57 & \underline{0.809} & 0.169 & \underline{9.59} \\
     && Feature-level fusion (Ours) & \textbf{7.56} & \textbf{0.818} & \underline{0.389} & \textbf{9.28} \\
    \bottomrule
  \end{tabular}
\end{table*}

\paragraph{Ex. 3: Optimization of Feature Fusion Strategy}

Table \ref{tab:fusion} compares feature-level fusion strategies to determine the optimal integration of the extracted feature vectors from the three fingers: simple concatenation ($\mathbf{f}_{concat}$), averaging ($\bar{f}$), and the proposed hybrid fusion ($\mathbf{f}_{fusion}$).
The optimal strategy varies slightly across models; for instance, averaging (Avg) and simple concatenation (Concat) exhibited the lowest MAE for DenseNet-161 and ResNeXt-101, respectively.
However, simple concatenation is sensitive to local noise in individual fingers, whereas averaging may smooth out finger-specific age-related vascular changes.
In contrast, the proposed hybrid fusion (Concat+Avg) not only achieved the highest accuracy for DenseNet-121 and ResNeXt-50 but also maintained the correlation coefficient (Corr.) and CS@5 at consistently high levels across all models.
This result suggests that simultaneously referencing the unique structural changes of each finger (Concat) and the individual-level systemic aging biases (Avg) provides the most robust approach in terms of generalization performance.
Consequently, we adopt Concat+Avg as the fusion module in MAGE-Vein.

\begin{table*}[t]
  \centering
  \caption{Comparison of feature fusion strategies. Best results for each model are in bold, and second-best are underlined.}
  \label{tab:fusion}
  \begin{tabular}{llcccc}
    \toprule
    Backbone & Fusion Strategy & MAE [y/o] $\downarrow$ & Corr. $\uparrow$ & CS@5 $\uparrow$ & Std. $\downarrow$ \\
    \midrule
    \multirow{3}{*}{DenseNet-121 \cite{Hung-CVPR-2017}}
     & Concat & 7.71 & \underline{0.858} & 0.336 & \underline{8.44} \\
     & Avg    & \underline{7.00} & 0.848 & \underline{0.436} & 8.64 \\
     & Concat+Avg (Ours) & \textbf{6.76} & \textbf{0.864} & \textbf{0.483} & \textbf{8.25} \\
    \midrule
    \multirow{3}{*}{DenseNet-161 \cite{Hung-CVPR-2017}}
     & Concat & 7.32 & 0.851 & 0.406 & 8.49 \\
     & Avg    & \textbf{6.68} & \textbf{0.861} & \textbf{0.440} & \textbf{8.28} \\
     & Concat+Avg (Ours) & \underline{7.25} & 0.840 & \underline{0.415} & 8.83 \\
    \midrule
    \multirow{3}{*}{ResNeXt-50-32x4d \cite{Xie-cvpr-2017}}
     & Concat & \underline{7.61} & \underline{0.822} & \underline{0.360} & \underline{9.19} \\
     & Avg    & 7.91 & 0.818 & 0.330 & 9.29 \\
     & Concat+Avg (Ours) & \textbf{7.40} & \textbf{0.823} & \textbf{0.402} & \textbf{9.14} \\
    \midrule
    \multirow{3}{*}{ResNeXt-101-32x8d \cite{Xie-cvpr-2017}}
     & Concat & \textbf{7.30} & \textbf{0.821} & \textbf{0.436} & \textbf{9.19} \\
     & Avg    & 7.78 & 0.801 & \underline{0.392} & 9.62 \\
     & Concat+Avg (Ours) & \underline{7.56} & \underline{0.818} & 0.389 & \underline{9.28} \\
    \bottomrule
  \end{tabular}
\end{table*}

\paragraph{Ex. 4: Impact of Multi-Task Learning}

Finally, based on the hypothesis that gender acts as a confounding factor in age estimation, we evaluate the effectiveness of multi-task learning and conduct a sensitivity analysis of the hyperparameter $\lambda$.
Table \ref{tab:lamda} shows the changes in age estimation accuracy for each backbone as $\lambda$ varies from 0 (single-task) to 100.
By selecting an appropriate $\lambda$, multi-task learning exhibited higher accuracy than single-task age estimation (Ex. 3) across all backbones.
Notably, with DenseNet-161 and $\lambda=20$, the MAE improved to 6.47, achieving the highest correlation (Corr. 0.876).
This improvement suggests that optimizing the gender classification task ($L_{CE}$) allows the network to condition on gender-specific traits, such as vessel thickness and hemoglobin concentration, thereby enabling the age regression head ($L_{MSE}$) to separate and focus on structural aging signatures.
However, large values (e.g., $\lambda=30$ for DenseNet-161) cause the gender classification gradient to become dominant, leading to a sharp decline in age estimation performance.

These four ablation studies identify the optimal configuration for MAGE-Vein: a DenseNet-161 backbone, padding preprocessing, feature-level fusion (Concat+Avg), and multi-task learning with $\lambda=20$.
This configuration is employed as the final proposed framework for subsequent comparative experiments.

\begin{table}[t]
  \centering
  \small
  \caption{Sensitivity analysis of the hyperparameter $\lambda$ in multi-task learning.}
  \label{tab:lamda}
  \resizebox{\linewidth}{!}{%
  \begin{tabular}{lccccc}
    \toprule
    Backbone & $\lambda$ & MAE [y/o] $\downarrow$ & Corr. $\uparrow$ & CS@5 $\uparrow$ & Std. $\downarrow$ \\
    \midrule
    \multirow{11}{*}{\shortstack[l]{DenseNet-121\\ \cite{Hung-CVPR-2017}}} 
     & 0  & 6.76 & 0.864 & \underline{0.483} & 8.25 \\
     & 1  & 8.04 & 0.861 & 0.416 & 8.57 \\
     & 10 & 7.62 & 0.834 & 0.455 & 8.99 \\
     & 20 & 8.46 & 0.854 & 0.439 & 8.87 \\
     & 30 & 8.41 & \textbf{0.870} & 0.413 & 8.38 \\
     & 40 & 7.06 & \textbf{0.870} & 0.474 & \textbf{8.19} \\
     & 50 & 6.63 & 0.855 & 0.455 & 8.31 \\
     & 60 & 7.70 & \underline{0.867} & 0.327 & 8.50 \\
     & 70 & 7.30 & 0.852 & 0.451 & 8.73 \\
     & 80 & \underline{6.49} & 0.862 & \textbf{0.504} & 8.35 \\
     & 90 & \textbf{6.48} & 0.862 & 0.471 & \underline{8.24} \\
     & 100& 11.59 & 0.820 & 0.307 & 9.73 \\
    \midrule
    \multirow{11}{*}{\shortstack[l]{DenseNet-161\\ \cite{Hung-CVPR-2017}}} 
     & 0  & 7.25 & 0.840 & 0.415 & 8.83 \\
     & 1  & 7.09 & 0.838 & \textbf{0.458} & 8.83 \\
     & 10 & 7.57 & 0.830 & 0.407 & 9.24 \\
     & 20 & \textbf{6.47} & \textbf{0.876} & \underline{0.455} & \textbf{8.05} \\
     & 30 & 11.01 & 0.851 & 0.289 & 8.82 \\
     & 40 & 8.88 & 0.866 & 0.308 & \underline{8.25} \\
     & 50 & \underline{6.63} & 0.865 & 0.455 & 8.31 \\
     & 60 & 7.30 & 0.854 & 0.387 & 8.76 \\
     & 70 & 7.49 & 0.856 & 0.333 & 8.67 \\
     & 80 & 8.19 & 0.857 & 0.295 & 8.76 \\
     & 90 & 8.26 & 0.838 & 0.433 & 9.07 \\
     & 100& 7.50 & 0.857 & 0.357 & 8.62 \\
    \midrule
    \multirow{11}{*}{\shortstack[l]{ResNeXt-50\\ -32x4d \cite{Xie-cvpr-2017}}} 
     & 0  & 7.40 & 0.823 & 0.402 & 9.14\\
     & 1  & 7.70 & 0.821 & 0.356 & 9.39 \\
     & 10 & 8.30 & 0.821 & 0.422 & 9.33 \\
     & 20 & \underline{6.93} & \textbf{0.857} & 0.418 & \underline{8.44} \\
     & 30 & 7.50 & 0.836 & 0.354 & 8.95 \\
     & 40 & 7.17 & 0.846 & 0.414 & 8.76 \\
     & 50 & 7.77 & 0.826 & \underline{0.428} & 9.22 \\
     & 60 & 8.22 & 0.788 & 0.382 & 9.93 \\
     & 70 & \textbf{6.79} & \underline{0.856} & \textbf{0.450} & \textbf{8.42} \\
     & 80 & 7.01 & 0.854 & 0.406 & 8.60 \\
     & 90 & 7.30 & 0.843 & 0.368 & 8.72 \\
     & 100& 7.51 & 0.827 & 0.367 & 9.15 \\
    \midrule
    \multirow{11}{*}{\shortstack[l]{ResNeXt-101\\ -32x8d \cite{Xie-cvpr-2017}}} 
     & 0  & 7.56 & 0.818 & 0.389 & 9.28\\
     & 1  & 8.79 & 0.804 & 0.283 & 9.99 \\
     & 10 & 7.74 & 0.810 & 0.375 & 9.45 \\
     & 20 & \underline{6.93} & \underline{0.857} & \underline{0.418} & \underline{8.44} \\
     & 30 & 10.72 & 0.770 & 0.222 & 10.48 \\
     & 40 & 7.47 & 0.811 & 0.392 & 9.45 \\
     & 50 & 7.61 & 0.818 & 0.360 & 9.30 \\
     & 60 & 7.80 & 0.812 & 0.328 & 9.42 \\
     & 70 & 9.02 & 0.805 & 0.357 & 9.63 \\
     & 80 & 8.71 & 0.828 & 0.398 & 9.18 \\
     & 90 & 8.83 & 0.834 & 0.284 & 9.29 \\
     & 100& \textbf{6.54} & \textbf{0.862} & \textbf{0.467} & \textbf{8.33} \\
    \bottomrule
  \end{tabular}
  }
\end{table}

\subsection{Comparison with Existing Methods}

This section evaluates the performance of MAGE-Vein.
Specifically, using the final model trained with data augmentation, we compare its performance with existing methods in both age and gender estimation tasks and verify its generalization performance on a public dataset.

\paragraph{Age Estimation Performance}

\begin{figure}[t]
  \centering
  \includegraphics[width=.9\linewidth]{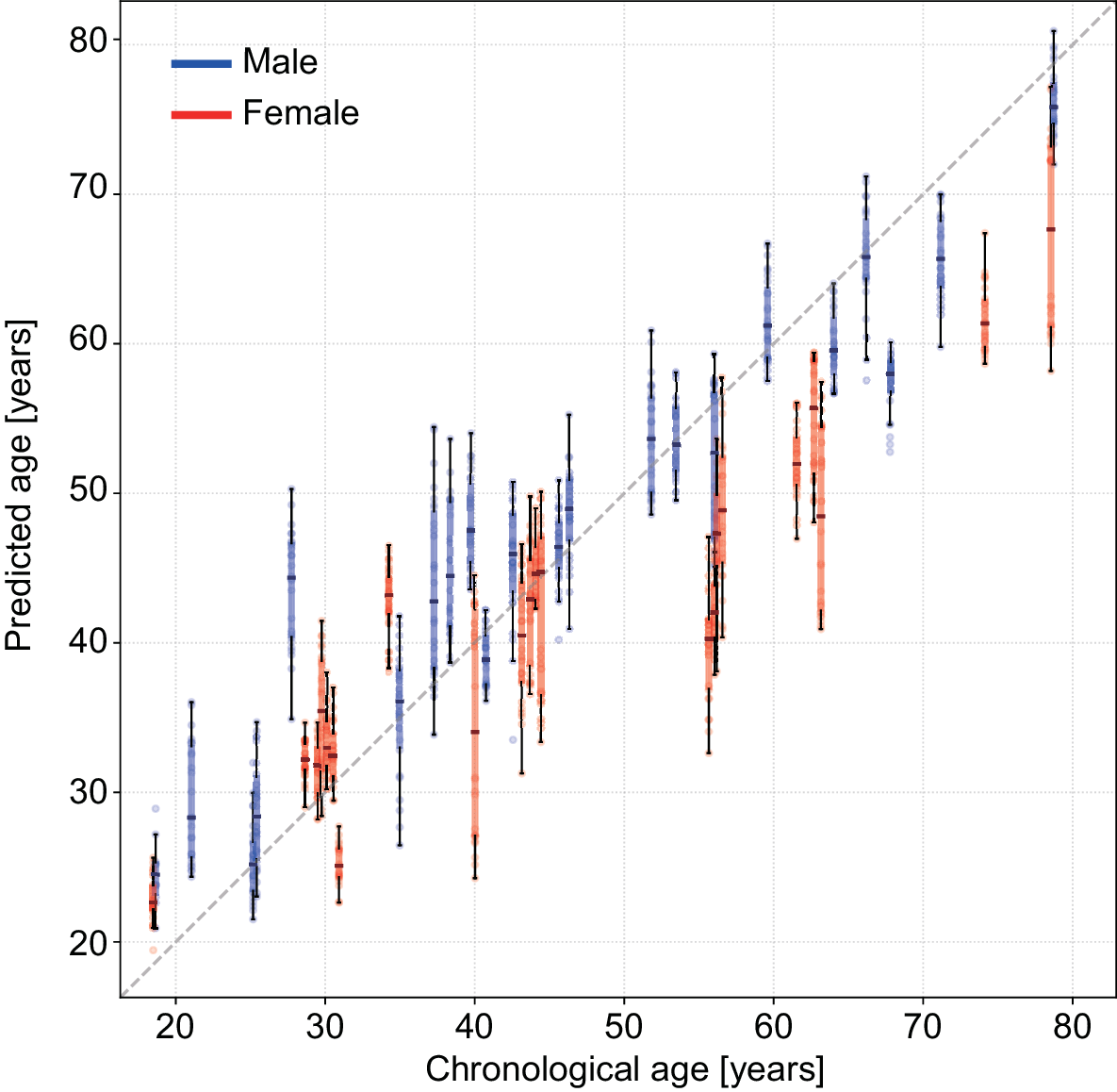}
  \caption{Boxplots showing the distributions of chronological and estimated ages per subject (blue: male, red: female).}
  \label{fig:plot}
\end{figure}

Table \ref{tab:augmentation_comparison} compares the age estimation accuracy among the proposed method, a baseline that uniformly outputs the mean age of the dataset (Dataset Avg.), and an existing method that attempted age regression from finger veins (Wimmer et al. \cite{Wimmer-ICPRW-2023}).
Compared to the existing method, the proposed method with data augmentation (Proposed w/ Aug.) achieved a significant improvement, yielding an MAE of 6.12 and a correlation of 0.880.
This result demonstrates the feasibility of accurate age estimation from finger vein images, challenging the previous negative conclusions.
Fig. \ref{fig:plot} shows the distribution of chronological and estimated ages for each subject in the test set.
To visualize the variance of the estimations across multiple samples from the same individual, the results are plotted using boxplots for each subject.
The entire cluster is distributed along the diagonal line, indicating that the model successfully avoids overfitting to the mean age and maintains linearity across a broad range of age groups.
Although individual vascular differences can cause certain subjects to be estimated as younger or older than their chronological age, the medians of the estimations for the same individual generally converge near their true age.
Considering the deployment in actual biometric systems, we also evaluated the estimation error at the subject level rather than the image level.
When the average of the estimated values from all test samples of a subject was used as their final estimated age, the MAE improved to 5.47, and the correlation increased to 0.921 (CS@5 of 0.523, Std. of 6.85).
This result demonstrates that local image-level errors caused by finger placement or imaging conditions are effectively canceled out by aggregating multiple samples, thereby realizing robust and reliable age estimation at the subject level.

\begin{table}[t]
  \centering
  \caption{Comparison of age estimation accuracy.}
  \label{tab:augmentation_comparison}
  \resizebox{\linewidth}{!}{%
    \begin{tabular}{lcccc}
      \toprule
      Model & MAE [y/o] $\downarrow$ & Corr. $\uparrow$ & CS@5 $\uparrow$ & Std. $\downarrow$ \\
      \midrule
      Dataset Avg. & 13.61 & -- & 0.184 & 16.08 \\
      Wimmer et al. \cite{Wimmer-ICPRW-2023} & 9.33 & 0.700 & 0.331 & 11.59 \\
      \midrule
      Proposed & \underline{6.47} & \underline{0.876} & \underline{0.455} & \underline{8.05} \\
      Proposed w/ Aug. & \textbf{6.12} & \textbf{0.880} & \textbf{0.526} & \textbf{7.77} \\
      \bottomrule
    \end{tabular}
  }
\end{table}

\paragraph{Gender Estimation Performance}

We evaluate the performance of gender classification, a secondary task in MAGE-Vein.
Table \ref{tab:gender} compares the F1-scores of the proposed method with state-of-the-art methods targeting gender estimation from finger vein images \cite{Wimmer-ICPRW-2023,Kuzu-Access-2023}.
Although designed primarily for age estimation, the proposed method achieved a total F1-score of 0.836, which is higher than those of methods specifically optimized for gender classification, such as Kuzu et al. \cite{Kuzu-Access-2023} (0.822) and the DenseNet-based method by Wimmer et al. \cite{Wimmer-ICPRW-2023} (0.829).
Notably, the proposed method achieved a F1-score of 0.857 in female classification.
This performance improvement results not only from the effect of multi-task learning but also from the proposed feature-level fusion.
Gender differences in finger vein patterns are obscured by local noise, such as variations in finger pressure or contact angles, when using a single finger.
By integrating features from three fingers, the network can robustly extract gender-specific vascular traits common to the individual, thereby enabling accurate classification.

\begin{table}[t]
  \centering
  \caption{Comparison of gender classification performance (F1-score $\uparrow$).}
  \label{tab:gender}
  \resizebox{\linewidth}{!}{%
    \begin{tabular}{lccc}
      \toprule
      Method & Male & Female & Total \\
      \midrule
      Wimmer et al. (DenseNet) \cite{Wimmer-ICPRW-2023} & \underline{0.822} & \underline{0.830} & \underline{0.829} \\
      Wimmer et al. (SqueezeNet) \cite{Wimmer-ICPRW-2023} & 0.769 & 0.774 & 0.771 \\
      Kuzu et al. \cite{Kuzu-Access-2023} & \textbf{0.824} & 0.821 & 0.822 \\
      \midrule
      Proposed (MAGE-Vein) & 0.815 & \textbf{0.857} & \textbf{0.836} \\
      \bottomrule
    \end{tabular}
  }
\end{table}
\begin{table}[t]
  \centering
  \caption{Results of 5-fold cross-validation on MMCBNU\_6000 \cite{Lu-CISP-2013}.}
  \label{tab:crossval}
  \resizebox{\linewidth}{!}{
  \begin{tabular}{llcccccc}
    \toprule
    \multirow{2}{*}{Metric} & \multirow{2}{*}{Method} & \multicolumn{5}{c}{Fold} & \multirow{2}{*}{Avg.} \\
    \cmidrule(lr){3-7}
    & & 0 & 1 & 2 & 3 & 4 & \\
    \midrule
    \multirow{3}{*}{\shortstack[l]{MAE\\ $\downarrow$}} 
     & Dataset Avg. & \underline{4.03} & \textbf{5.40} & \textbf{5.50} & 3.84 & \textbf{4.57} & \underline{4.67} \\
     & Wimmer et al. & 4.37 & \underline{5.51} & \underline{5.63} & \underline{3.78} & 5.00 & 4.86 \\
     & Proposed & \textbf{3.82} & 5.67 & 5.66 & \textbf{3.37} & \underline{4.99} & \textbf{4.63} \\
    \midrule
    \multirow{3}{*}{\shortstack[l]{Corr.\\ $\uparrow$}} 
     & Dataset Avg. & -- & -- & -- & -- & -- & -- \\
     & Wimmer et al. & \underline{0.022} & \underline{0.019} & \underline{0.176} & \underline{0.333} & \textbf{0.129} & \underline{0.139} \\
     & Proposed & \textbf{0.108} & \textbf{0.160} & \textbf{0.310} & \textbf{0.567} & \underline{0.083} & \textbf{0.286} \\
    \midrule
    \multirow{3}{*}{\shortstack[l]{CS@5\\ $\uparrow$}} 
     & Dataset Avg. & \textbf{0.750} & \underline{0.700} & 0.650 & \textbf{0.900} & \textbf{0.850} & \textbf{0.770} \\
     & Wimmer et al. & 0.690 & \textbf{0.753} & \textbf{0.681} & 0.844 & 0.724 & \underline{0.739} \\
     & Proposed & \underline{0.700} & 0.688 & \underline{0.655} & \underline{0.895} & \underline{0.730} & 0.734 \\
    \midrule
    \multirow{3}{*}{\shortstack[l]{Std.\\ $\downarrow$}} 
     & Dataset Avg. & \textbf{4.31} & \underline{10.59} & 8.68 & 7.26 & \textbf{7.88} & \underline{7.75} \\
     & Wimmer et al. & 5.33 & 10.97 & \underline{8.56} & \underline{6.89} & 8.16 & 7.98 \\
     & Proposed & \underline{4.66} & \textbf{10.56} & \textbf{8.25} & \textbf{6.27} & \underline{8.04} & \textbf{7.56} \\
    \bottomrule
  \end{tabular}
  }
\end{table}

\paragraph{Evaluation on Public Dataset}

To compensate for the lack of public availability of our dataset and evaluate generalization, we conducted a 5-fold cross-validation on MMCBNU\_6000 \cite{Lu-CISP-2013}, the public dataset used by Wimmer et al. \cite{Wimmer-ICPRW-2023}.
Although MMCBNU provides three-finger images with age labels, its demographic distribution is heavily skewed toward the 20s and 30s as mentioned in Sect. \ref{sec:related_work}.
As shown in Table \ref{tab:crossval}, while MAGE-Vein achieved a lower MAE (4.63) than the method by Wimmer et al. (4.86), all methods exhibited an abnormal discrepancy: extremely low correlation coefficients (0.1--0.2) despite low MAEs around 4 years.
In fact, the baseline predicting only the dataset's mean age (Dataset Avg.) achieved an MAE of 4.67, which is lower than that of Wimmer et al.
This coexistence of a low MAE and a low correlation demonstrates that the networks merely overfitted to the skewed dataset mean rather than learning age-related vascular changes.
These findings suggest that the previous negative conclusion regarding age estimation was an artifact primarily caused by the demographic bias of the dataset, rather than a limitation of the finger vein modality.
Simultaneously, this result justifies the necessity of the demographically balanced dataset constructed in this study.

\begin{figure}[t]
  \centering
  \includegraphics[width=.9\linewidth]{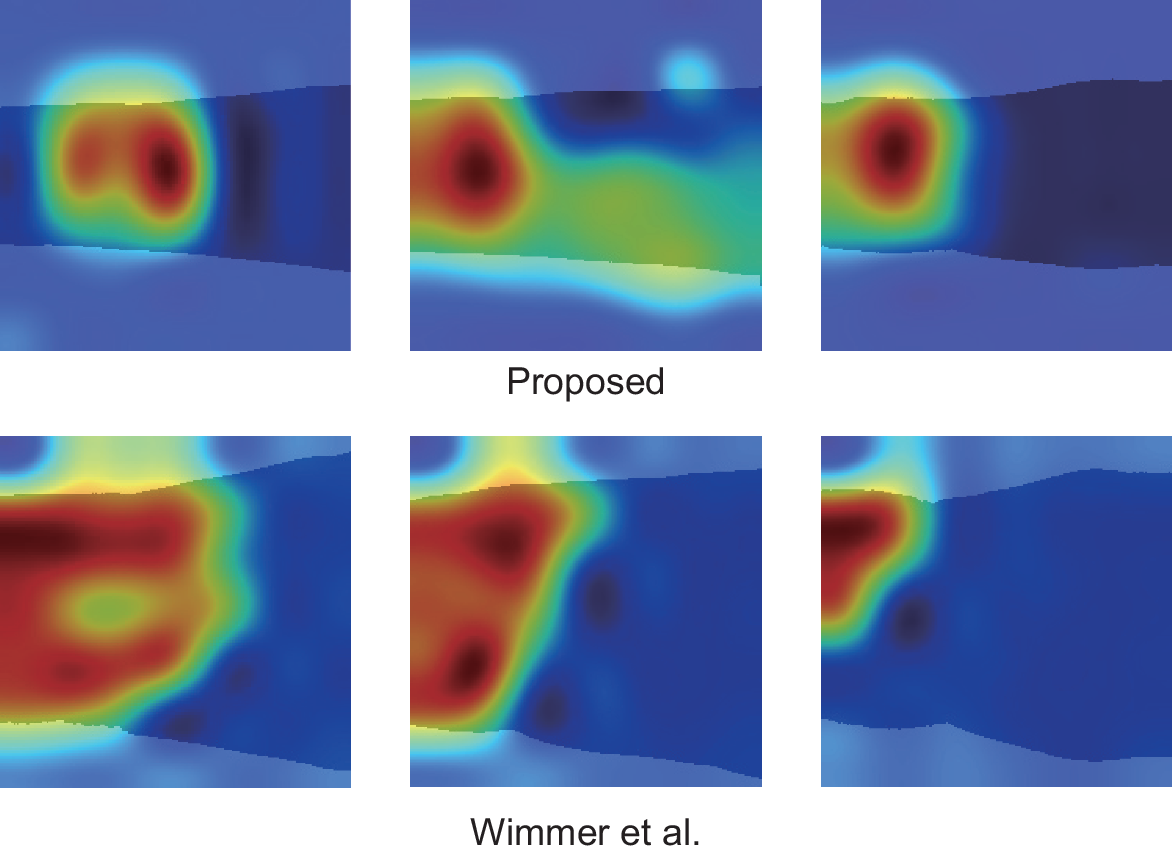}
  \caption{Comparison of attention maps generated by Grad-CAM \cite{Selvaraju-ICCV-2017}.}
  \label{fig:gradcam}
\end{figure}

\subsection{Visual Explanations}

To qualitatively evaluate the validity of methods, we visualize the regions contributing to age estimation by mapping gradients from the final convolutional layer using Grad-CAM \cite{Selvaraju-ICCV-2017}.
For privacy protection, the heatmaps are overlaid on binarized finger silhouettes rather than raw vein images.
Fig. \ref{fig:gradcam} compares the attention maps of the conventional method \cite{Wimmer-ICPRW-2023} and MAGE-Vein for the same subject.
Note the difference in aspect ratios due to their preprocessing methods, i.e., resizing and padding.
The conventional method exhibits strong activations along the finger contours and background boundaries.
This result indicates overfitting to the anatomical distortion caused by resizing and global shape noise.
Conversely, the proposed method, which combines padding and multi-task learning, effectively suppresses these boundary activations and focuses its attention on the internal finger regions.
This visualization demonstrates the physiological validity of MAGE-Vein, confirming that it captures structural vascular changes without being misled by imaging noise.

\section{Conclusion}

This paper proposed MAGE-Vein, a multi-instance, multi-task learning framework designed for accurate age and gender estimation from finger vein images.
By employing a hybrid fusion strategy that combines the concatenation and averaging of features extracted from three fingers of the same subject, the proposed method suppresses local noise and robustly extracts universal age-related changes common to the individual.
Furthermore, simultaneously optimizing the gender classification task effectively eliminates the confounding effects of gender-specific vascular traits, thereby improving the age estimation accuracy.
Experimental evaluations using a demographically balanced large-scale dataset and a public dataset demonstrated that the proposed method achieves highly accurate age estimation.
This result challenges the previous negative conclusions that age estimation from this modality is impractical.
In future work, we plan to leverage the attribute information obtained by this method to develop a highly accurate finger vein authentication system that is robust to aging effects.

\section{Acknowledgment}

This work was supported in part by JSPS KAKENHI 23H00463, 23H03395, and 25K03131.

{\small
\bibliographystyle{ieee}
\bibliography{egbib}
}

\end{document}